\documentclass[journal]{IEEEtran}

\usepackage{cite}
\usepackage{amssymb,amsfonts}
\usepackage{graphicx}
\usepackage{textcomp}
\usepackage{xcolor}
\usepackage{algorithm,algorithmic,amsmath}

\include{pythonlisting}

\begin{document}

\title{Prune2Edge: A Multi-Phase Pruning Pipelines to Deep Ensemble Learning in IIoT}

\author{Besher Alhalabi$^1$, Mohamed Medhat Gaber$^1$ and Shadi Basurra$^1$ 
\thanks{$^1$Authors are with the School of Computing \& Digital Technology, Birmingham City University, United Kingdom {\tt\small besher.alhalabi@bcu.ac.uk},        {\tt\small mohamed.gaber@bcu.ac.uk},    {\tt\small shadi.basurra@bcu.ac.uk}
}}

\markboth{a revised version is going to be submitted to a journal soon}%
{Shell \MakeLowercase{\textit{et al.}}: }

\maketitle

\begin{abstract}
Most recently, with the proliferation of IoT devices, computational nodes in manufacturing systems (Industrial-Internet-of-things), and the lunch of 5G networks, there will be thousands of connected devices generating a massive amount of data. In such an environment, the controlling systems need to be intelligent enough to deal with a vast amount of data to detect defects in a real-time process. Driven by such a need, artificial intelligence models such as deep learning have to be deployed into IIoT systems. However, learning and using deep learning models are computationally expensive, so an IoT device with limited computational power could not run such models. To tackle this issue, edge intelligence had emerged as a new paradigm towards running Artificial Intelligence models on edge devices. Although a considerable amount of studies have been proposed in this area, the research is still in the early stages.
In this paper, we propose a novel edge-based multi-phase pruning approach to ensemble learning on IIoT devices. First, we generate a diverse ensemble of pruned models, then we apply integer quantisation, next we prune the generated ensemble using a clustering-based technique. Finally, we choose the best representative from each generated cluster to be deployed to a distributed IoT environment. On CIFAR-100 and CIFAR-10, our proposed approach was able to outperform the predictability levels of a baseline model (up to 7\%), more importantly, the generated learners have a relatively small sizes (up to 90\% reduction in the model size) that minimise the required computational capabilities to make an inference on the resource-constraint devices.

\end{abstract}

\begin{IEEEkeywords}
IIoT, IoT, resource-limited environments, deep learning, edge-ai, quantisation, weight pruning, ensemble pruning, ensemble learning.
\end{IEEEkeywords}

\IEEEpeerreviewmaketitle

\section{Introduction}
We currently witness the rise of the fourth industrial revolution known as Industrial Internet Of Things (IIoT) \cite{xu_survey_2018}. In short, IIoT is the utilisation of IoT devices in assembly lines and manufacturing processes, so the automation of monitoring and control of manufacturing systems could be achieved. The key factor in such monitoring systems is to provide intelligence through IoT generated big data to detect any defects during manufacturing. Thus, IoT devices in industrial domains are regarded as an optimal target for Artificial Intelligence (AI) applications as they are continuously generating a massive amount of data, and machine learning algorithms typically need to be fed by large data sets to produce accurate models. However, the limitation of the computational power for IoT devices prevents running advanced AI models like deep learning. Typically, the state of the art deep learning models have millions of trainable parameters, and they require extremely powerful workstations to be trained. For example, VGG-16 requires around 550 MB of memory and has almost 138 million parameters \cite{simonyan_very_2014}. As a result, it is very challenging to train or deploy deep learning model on resource-limited devices. To tackle all preceding issues, the research community addressed the following:\\
1. \emph{The complexity of deep neural network}: To overcome the complexity of deep neural networks and allow them to be deployed on resource-limited devices, many compression and acceleration strategies have been introduced by the deep learning community including (1) simple regularisers as ($L1$ and $L2$) that are commonly used during the training phase to control the complexity of a neural network \cite{nowlan_simplifying_1992} \cite{girosi_regularization_1995}. DropConnect also could be used to prune a network by randomly dropping a subset of weights \cite{wan_regularization_2013}; (2) neuron pruning and sharing methods that increase the sparsity of neural networks by removing irrelevant connections \cite{sietsma_neural_1988}\cite{hassibi_optimal_1993}\cite{hong-jie_xing_two-phase_2009}\cite{zhang_deep_2018} \cite{hagiwara_simple_1994} \cite{han_deep_2015}. Network quantisation could fit under this category as it aims to reduces the number of required bits to represent the network's weights \cite{gong_compressing_2014}\cite{wu_quantized_2016}; (3) knowledge distillation aims to transfer the knowledge from a teacher model (large network) into a student model (small network) \cite{romero_fitnets:_2014}\cite{you_learning_2017}\cite{yim_gift_2017}; (4) compact network design that works toward reducing the complexity and improving the accuracy of the whole neural network. This is achieved by the use of optimisation of the network's architecture and storage \cite{chollet_xception:_2016}\cite{howard_mobilenets:_2017}\cite{howard_searching_2019} \cite{Kamada2018AdaptiveSL}. However, Only a few of the previous methods were able to produce compressed models without a significant drop in the model's performance.\\
2. \emph{Limitation of the computational capabilities of IoT devices}: due to the low memory and processing power of the of those devices, the computation could be moved to be performed on a powerful remote workstation, typically cloud servers \cite{satyanarayanan_emergence_2017}. However, this approach makes no attempt to consider communication costs, network latency and data privacy. Edge computing has emerged as a promising solution for the previous challenges, instead of offloading the data to a remote infrastructure, an edge computing device is going to be attached close to the resource-limited devices where the data is being generated. For instance, a Google coral device could be added to a Raspberry Pi attached to a camera to do object detection using the state of the art deep learning models. One possible approach to utilise deep learning models on IoT is to divide the neural network into two parts, one part will be deployed to the edge and the other part will be deployed on the cloud server \cite{osia_private_2018}. Such approach could reduce the communications costs and provide the same accuracy of the cloud-based approach, but it is hard to combine the data from two different neural networks because different networks have different parameters and computational overhead. Although those techniques could significantly reduce the network size, it could affect the network's performance due to compression.

In this paper, we propose a novel edge-based deep pruning approach powered by the synthesis of deep learning ensembles \cite{alhalabi_ensyth:_2019} which eliminates the high variance of deep learning models and produces outperforming classifiers in term of accuracy, generalisation and inference time. Motivated by the lottery ticket hypothesis \cite{frankle_lottery_2019}, this approach starts by generating a diverse set of pruned deep learning models using different hyperparameters of the pruning algorithms, aiming at finding a diverse ensemble of subnetworks, a coalition of lottery tickets, instead of one to be deployed on a resource limited environment.
Next, integer quantisation is applied to all pruned models to ensure maximum performance when they get deployed on AI edge endpoints that include TPUs or CPUs. Then, we apply the clustering-based ensemble pruning technique to select a subset of the classifiers from the generated pool. Based on the generated clusters, we elect from each cluster a few representatives based on two proposed strategies: accuracy first or diversity first. After that, we deploy the elected models to a distributed edge environment where each node could empower ensemble learning and combine the predictions from two or more classifiers to generate high confidence predictions. Experimentally, on CIFAR-10, CIFAR-100 our proposed method was able to produce classifiers with higher accuracy levels with up to 90\% reduction in model size.

The rest of this work is organised as follows. In Section II, we introduce some of the most recent related work in the field of accelerating deep neural networks and mapping them on the edge. In Section III, we explain in details our proposed deep-pruning approach for deep learning models toward efficient deployment on resource-limited systems. Next in Section IV, we discuss the experiments on CIFAR10 and CIFAR100, then we present the deployment of our outperforming classifiers into a distributed edge environment. In the same section, we provide some bench-marking results related to energy consumption, heat and inference time to prove the applicability of such an approach in IIoT. Finally, we conclude this paper and present future work in Section V.

\section{Related Work}
\label{related_work}
With the breakthroughs for deep learning applications and services in a wide range of domains, the research community have shown an increased interest in edge intelligence to unleash the potential of neural networks by learning from the edge data. However, due to the limited computational resources for network devices like IIoT, the deep neural network could be neither trained nor deployed directly on edge. A considerable amount of literature has been published proposing different architectures, technologies and key performance indicators to enable efficient distributed training for edge intelligence models. In the following subsections, we list some existing architectures, and the most recent methods that enable this particular way of model's training.

\subsection{Architectures}
\begin{figure}[htbp]
    \centering
    \includegraphics[scale=0.6]{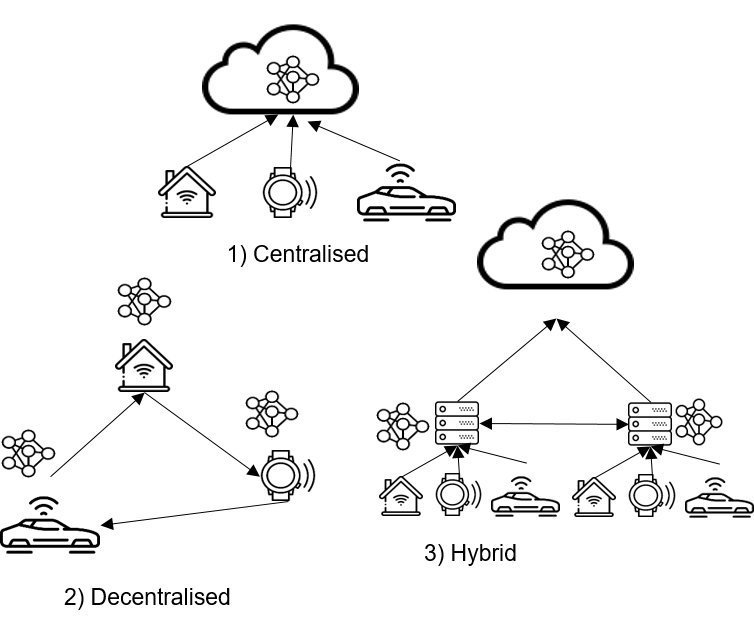}
    \caption{Distributed training's architectures}
    \label{parallel-training-arch}
\end{figure}

Fig.\ref{parallel-training-arch} shows three different architectures for distributed deep neural networks training at the edge \cite{zhou_edge_2019}. \textbf{1)Centralised}: in this approach, the DNN model is trained in a cloud server as follows:  training's data is collected from edge devices (mobiles, smart house appliances, sensors, etc.) then it will be sent to a cloud server where a DNN model is deployed. The model then will use the data for training. \textbf{2)Decentralised}: each edge device in this mode (node) trains its own DNN model based on the local data preserving local privacy. Next, all nodes could collaborate to produce a shared model by exchanging model's update. \textbf{3) Hybrid}: as shown in Fig.\ref{parallel-training-arch} hybrid architectures combined both centralised and decentralised modes. The DNN model could be trained on either a cloud server or by receiving decentralised updates from the nodes. 
\subsection{Methods} in the following we review the most recent methods for enabling distributed DNN training on the edge.
   \textbf{a)federated learning}: federated learning is emerging as promising solution for parallel deep learning training. Unlike the traditional training approaches where the data has to be sent to a central server, federated learning \cite{mcmahan_communication-efficient_2017} \cite{chen_data-driven_2019} allows all nodes (IoT devices) to train their own models based on local data then train a shared model in a central server by aggregating smart updates from the edge nodes. \cite{shokri_privacy-preserving_2015} proposes a new protocol SSGD (Selective Stochastic Gradient Descent) that allows the edge nodes to train on the local data sets, and selectively share a subset of their parameters to be sent to the centralised server. \cite{kim_blockchained_2019} suggests to reply on blokchain technique to verify the distributed models updates after transmitting them to the central aggregator and the main advantage of this approach that it could also works on decentralised architecture. For unreliable networks, federated learning could be very challenging to apply as the clients need to send frequent updates to the centralised model, \cite{konecny_federated_2017} found that increasing the computation of local updates on each client could help to solve this issue because the number of communications with the central server is going to be decreased. 
   
   \textbf{b)DNN architecture splitting}: in this approach the deep learning model is splitted into two partitions to be deployed between two different locations (one to be deployed on the edge device and the other on the edge server). However in a distributed training environment, it's a challenging task to determine the splitting point that preserves the model's accuracy and user's privacy. \cite{mao_privacy-preserving_nodate} employs the differentially private mechanism to split the DNN after the first convolutional layer between the edge server and the edge device in away that protect both of models parameters and private data. \cite{harlap_pipedream:_2018} proposes PipeDream a pipelined training framework that automatically finds the best way to split an input model across the available edge nodes. Pipedream minimises the communications (up to 95\% for large DNN models) and allows the optimal overlap of computation and communication. 
 
   \textbf{c)knowledge distillation}: KD (Knowledge distillation) is widely use to compress a deep neural network into a shallower network, it aims to transfers the knowledge that the weights had learned in the original network (teacher net) into the shallower network (student net). \cite{caruana_ensemble_2004} was able to transfer a compressed knowledge from an ensemble into a single model. \cite{hinton_distilling_2015} was able to extend the previous work to develop a new compression framework using KD. The framework compresses the teacher model (ensemble of deep neural nets) into a student model by training the student model to predict the output of the teacher in addition to the original classification labels.
    \textbf{d)Fine-tuned architecture}: This category involves optimisation for the network's architecture and the convolutional layers' design (if applicable). \cite{howard_mobilenets:_2017} uses depth-wise convolutions to generate compressed models that could easily fit on mobile devices. Depth-wise convolutions were firstly introduced by \cite{sifre_rigid-motion_2014}, then it was adopted by \cite{ioffe_batch_2015} to present the Inception model. Inception model primarily based on co-variate shifts which is a useful technique in minimising the number of activation and reduce the training time.

\section{\emph{Prune2Edge}} \label{method}
In this section, we describe in details the proposed approach. We start by justifying the rationale behind generating a pool of deep learning classifiers and introducing the pruning approach that had been applied to the pool. Then we explain the post-training quantisation that gives an advantage when it comes to deploying the deep learning models on the edge. After that, we move to illustrate how to reduce the number of the models in the generated pool by proposing a clustering-based ensemble pruning method and then how to choose representatives from each cluster.
\subsection{pool generation and weight pruning}
Deep learning models can solve none linear problems by learning via stochastic training algorithm. This makes them able to learn complex relationships between inputs and outputs and to approximate any mapping function. However, one major drawback is that deep learning models suffer from high variance which makes them highly dependent on: 1- the training set that used to train the model, 2- the conditions that been applied to the training process (initial weights values, loss function, optimiser function, etc.). This could affect the network's ability to generalise and produces a final model that makes different predictions when the same conditions apply. To overcome those challenges, we follow a similar approach applied here \cite{alhalabi_ensyth:_2019}. A diverse set of pruned models are going to be generated and will be considered as a pool. Weight pruning described in \cite{zhu_prune_2017} is going to be applied. During the training process, a binary mask is added to each elected layer for pruning, the mask will hold both same size and shape of the layer's tensor, and it will determine the weights that actively participates in the forward execution graph. Then the weights in each mask are going to be sorted according to their absolute values, and the ones with the smallest magnitude values are going to be set to zero, this will lead to initial sparsity levels ($s$). During the backpropagation, the weights that had been marked in the forward execution are not going to be updated. This process will be executed automatically and gradually where the sparsity is increased from initial value ($s_i$) (normally 0) to final sparsity ($s_f$) over several pruning steps ($n$), starting at a training step ($t_0$) with a pruning frequency ($\Delta t$):
\begin{equation}
    s_t = s_f + (s_i - s_f) {\bigg(1 - \frac{t - t_0}{n \Delta t} \bigg)}^3 
\end{equation}
    where $t \in \{ t_0, t_0 + \Delta t, ...,  t_0 + n \Delta t \}$

($s$,$n$,$t$,$ \Delta t$) are the hyperparameters for this pruning technique; we are going to assign different values from them and generate a pool of pruned deep learning models.
However, compressing the model's size using such a pruning technique is not adequate because the weights in the generated model are going to be represented by floating-point numbers. Dealing with floating-point may require powerful computational resources like GPUs that are not available in AI edge points. Thus, a post-integer quantisation needs to be applied in the next step.

\subsection{post training integer quantisation}
instead of representing the neural network's weights using floating-point numbers, 8-bit integer quantisation will approximate the floating-point values using the following formula \cite{noauthor_tensorflow_nodate}:
\begin{equation}
    real\_value = \left(int8\_value - zero\_point \right) \times scale 
\end{equation}
All weights are going to be represented by $int8$ two's complement values in range $[-127, 127]$ with $zero\_point$ equals to ($0$). Similarly, activations/inputs will be represented by $int8$ two's complement values in range $[-128, 127]$ with with $zero\_point$ in range $[-128, 127]$.
After applying integer quantisation to the proposed pool, we will get a set of pruned-quantised classifiers that have different characteristics. Some models could have excellent performance in terms of accuracy and inference time, while others do not. 
In the next step, we reduce the number of learners in the pool by applying ensemble pruning assuming the pool as an ensemble of deep learning models.
\subsection{ensemble pruning}
According to \cite{zhou_ensemble_2012} ensemble pruning methods could be mainly classified to: ordering based pruning, optimisation based pruning and clustering-based pruning. Typically, clustering-based techniques aim to partition the classifiers into different clusters where the classifiers that belong to the same cluster have common characteristics and behave similarly, while different clusters have different levels of diversity. Finding such clusters is critical to our approach because it will maximise the generalisation of the final solution and overcome any potential overfitting issues. Although there is a large volume of studies describing clustering-based ensemble pruning \cite{giacinto_design_2000} \cite{qiang_clustering-based_2005} \cite{lazarevic_effective_2001}, such experiments are not satisfactory nowadays because of the different nature of the modern computer vision bench-marking datasets that include images. We are proposing a new clustering-based ensemble pruning technique for deep learning ensembles that work on image datasets.

In order to produce high accurate ensembles for the final solution, we need to maintain
   1-accuracy of the ensemble: The accuracy of the ensemble could be improved by choosing well-trained individuals with high predictability levels, we do descending rank  to the learners after applying the ensemble pruning method based on the accuracy levels against a subset of the training set called the "pruning set", then we select representatives from each of the generated clusters. This will ensure that only high performing models are considered to be part of the final deep learning ensemble.
  
   2-diversity of the individual learners in the ensemble: although the `diversity` had been discussed by many researchers\cite{kuncheva_measures_2003}\cite{banfield_ensemble_2005}, there is still no standard definition for the diversity of an ensemble. Generally, the assumption is to have multiple learners with different values for prediction errors. So taking into account that the learners have been trained on the same dataset, we could assume that the higher levels of the differential in weights/output between the ensemble's members are going to bring higher diversity for the ensemble. Based on this, we could consider our proposed pool of classifiers as divers set, because the proposed models have been pruned with different values of hyperparmas that lead to different sparsity levels, weights and outputs.

Moving now to explain in details the proposed ensemble pruning method, let's assume  $S$ is pruning set (subset of the training set)$\in \mathbb{R}^{m \times n}$ where $m$ is the number of samples and $n$ is the number of labels. Let $m$ a deep learning model $\in P$ the  pool of pruned-quantised models, the prediction of the model $m_i \in P$ for a class $c_j \in S$ is $y_{i,j} \in [0,1] $ where:
\begin{itemize}
    \item $i = 1,2,\cdots,m$ where $m$ is the number of models.
    \item $j = 1,2,\cdots,n$ where $n$ is the  number of classes.
\end{itemize}
Now we could construct the output of the pool $P$ for one class $c_j$ as:
\begin{equation*}
A_j = 
\begin{bmatrix}
y_{1,1} & y_{1,2} & \cdots & y_{1,n} \\
y_{2,1} & y_{2,2} & \cdots & y_{2,n} \\
\vdots  & \vdots  & \ddots & \vdots  \\
y_{m,1} & y_{m,2} & \cdots & y_{m,n} 
\end{bmatrix}
\end{equation*}
where $j$ is the number of classes in the pool $P$. As the output of the pool has been defined, We move now to define how clustering is applied to pool's output $A_j$. Let's consider $x_j$ as a vector  $\in A_j$, the value of $x_j$ is defined as: 
$x_j = [A_b,j]$ where b 
\begin{equation*}
b = argmax
    \begin{bmatrix}
     max(A_{1,n}) \\
     max(A_{2,n}) \\
     \vdots  \\
     max(A_{m,n}
    \end{bmatrix}
\end{equation*}
Based on this a clustering matrix $C$ could be defined as : $C = [x_1 x_2 \cdots x_j]$, and the clustering results $R$ after applying kmeans is $ R= kmeans(C,k)$ where k is the number of clusters.
$R$ then is used to know the cluster for each model, the pool's models are clustered together based on their prediction's behaviour where different clusters have different levels of diversity and accuracy at the same time. Algorithm.~\ref{algo1} describes the complete ensemble pruning process.
\begin{algorithm}
  \begin{algorithmic}[1]
  \caption{clustering-based ensemble pruning}

    \FOR{$c_j \in S$}
     \STATE $calculate (A_J)$
     \STATE $b = argmax (max (A_J))$
     \STATE $x_j = [A_b,j]$
     $c \gets (x_j)$
    \ENDFOR
    \STATE $ R= kmeans(C,k)$
  \end{algorithmic}
  \label{algo1}
\end{algorithm}
Next step is to determine how to choose the best candidates from the resulting clusters to compose the final ensemble that will be deployed to the distributed constraint-limited environment.
\subsection{representative selection strategies}
we aim to deploy high-performing classifiers to the IoT devices and allow those tiny devices to employ ensemble learning to produce confident decisions. For this reason, we are proposing two strategies to select the representatives from the generated ensemble-pruning approach. 
\begin{itemize}
    \item accuracy first: in this approach, we aim to provide the maximum levels of accuracy. Thus we choose the cluster that holds the classifiers with the highest predictability levels taking into consideration that the classifiers were ranked in the previous step, then we select a few models to be deployed to the edge device.
    \item diversity first: for this strategy, we give priority to the diversity of the final solution,  the aim here is to provide different classifiers with higher levels of diversity to examine the effect when they are synthesised into one ensemble. we select one classifier from each of the generated clusters then we deploy those deep learning models to distributed constraint devices.
\end{itemize}
Next, the models are going to be deployed to a distributed constraint-limited environment; each node in this environment uses max voting \cite{zhou_ensemble_2012} as an ensemble learning method to synthesis the prediction of the deep learning models. The predictions of the models are considered as votes, then the class that receives the maximum number of votes is considered as the ensemble's prediction.
Let's consider $P= {m_1,\cdots,m_N}$ as an ensemble of the deployed models $m_i$, the prediction of the ensemble for a test class using max voting is the class that receives the maximum support $\eta_{final} (P)$ from the ensemble's members, based on this we could define the ensemble output as :
$$\eta_{final} (P) = argmax_{j\in \{1,\cdots,C\}}\sum_{i=1}^N y_{i,j} $$

\begin{figure}[htbp]
    \centering
    \includegraphics[ scale=0.3]{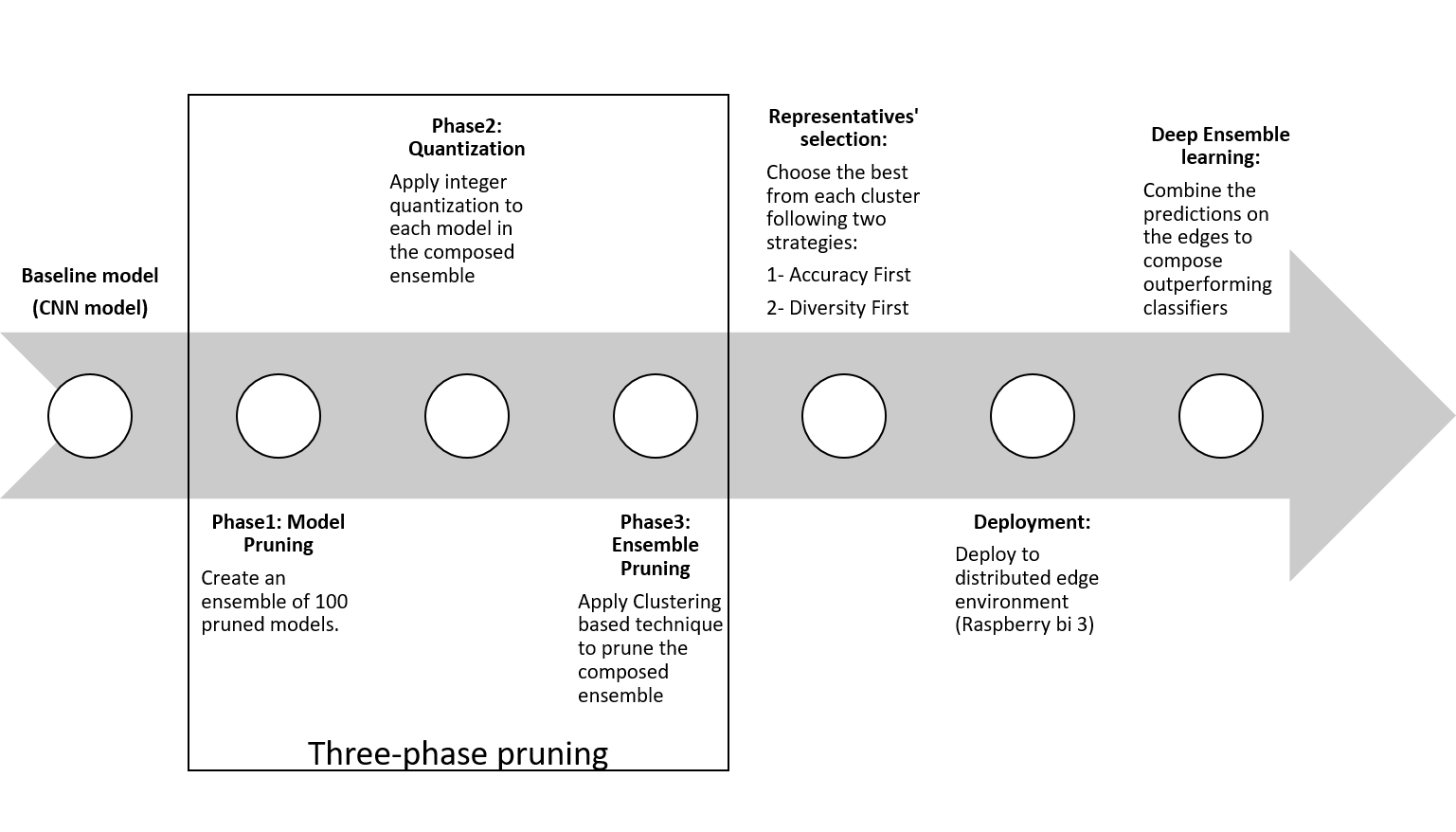}
    \caption{deep-pruning}
    \label{deep_pruning_method}
\end{figure}

Finally, as shown in Fig.~\ref{deep_pruning_method}, our approach to synthesis deep learning models on edge devices could be summarised as: 
\begin{itemize}
    \item train baseline deep learning model;
    \item create a diverse pool of pruned deep learning classifiers;
    \item convert the weights/outputs of the neural network to integer values by applying inter quantisation;
    \item prune the proposed pool adopting clustering-based technique;
    \item select the best models in the generated clusters;
    \item deploy the selected models to distributed edge environment;
    \item utilise ensemble learning to combine the predictions of the deployed models on the edge.
\end{itemize}
In the next section, we move to the experimental set up and the results of applying this method on a bench-marking datasets.

\section{Experiment}
in this section, the performance of \emph{Prune2Edge} is evaluated with CIFAR10CNN and ResNetV2 models on CIFAR10 and CIFAR100 datasets respectively.
Then, bench-marking results related to inference time, heat and power consumption levels are provided for running the \emph{Prune2Edge} models on Raspberry PI devices.

\subsection{Datasets and models}
\paragraph{
    \textbf{datasets}: \textbf{CIAFR-10} \cite{krizhevsky_learning_nodate} consists of 70,000 images (28x28 colour) splitted into 50,000 images as training set and 10,000 images as testing set. The images span over 10 objects' categories [0: airplane, 1: automobile, 2: bird, 3: cat, 4: deer, 5: dog, 6: frog, 7: house, 8: ship, 9: truck].
     \textbf{CIFAR-100} is identical to CIFAR10 except that it has 100 classes holding 600 images each.
}

\paragraph{
\textbf{models}: \textbf{CIFAR10-CNN} is basic convolutional neural network (CNN) model, it consists of four convolutional layers with (32*3*3) filter size for the first two and (64*3*3) filter size for the second. 
 \textbf{ResNetV2}: \cite{he_identity_2016} is the second version of ResNet. Mainly the improvements in V2 is related to the arrangement of layers in the residual block. The model's input is a 299×299 image, and the output is the probabilities distribution for the predicted class.
}
\subsection{Implementation details}
\subsubsection{model training and pruning}
CIFAR10-CNN model is trained from scratch on CIFAR10 dataset as a baseline model, the accuracy of the baseline model on the testing set is 74\% and the model size is 9,386 KB. Similarly, ResNetV2 model is trained on CIFAR100 dataset, and the accuracy of the baseline model reaches 67\% and the size of this model is 3,575 KB. Both baseline models are pruned using weight pruning \cite{zhu_prune_2017} to generate two pools of pruned deep learning models where each pool contains a hundred models from each baseline.
During the pruning process, different values for the pruning hyperparameters are provided to ensure maximum diversity between the models of the pools. The values are randomly chosen from a range of values, full list of the hyperparameters and the range of values are shown in Table.~\ref{table:hyperparams}

\begin{table}[ht]
\caption{hyperparameters, weight pruning}
\centering
\begin{tabular}{|p{2cm}| p{3.5cm}|} 
\hline
hyperparameter & values\_range  \\ 
\hline 
epochs & [3,4,5,6,8]  \\ 
\hline
batch\_size & [32,64,128]  \\ 
\hline
loss & [categorical\_crossentropy,
mean\_squared\_error,
mean\_absolute\_error]  \\ 
\hline
optimiser & [SGD,adam,Nadam,Adadelta]  \\ 
\hline
initial\_sparsity & [0.1,0.6](range)  \\ 
\hline
final\_sparsity & [0.7,0.9] (range)  \\ 
\hline
frequency & [100,200,300,400]  \\ 
\hline

\end{tabular}
\label{table:hyperparams} 
\end{table}

\subsubsection{integer quantisation}
After generating two pools of pruned deep learning models, we apply post-training integer quantisation on all models to convert activations and weights to 8-bit integers. Quantisation will lead to a significant reduction to the model's size and bring performance enhancements on integer-only hardware accelerators.

\subsubsection{ensemble pruning}
at this stage, two pools of diverse-pruned-quantised deep learning models are generated (100 models each). We consider each pool as an ensemble of individual learners, and we aim to choose the best performing models in each ensemble by applying our proposed ensemble pruning approach. After ensemble pruning, each ensemble will be splitted into several clusters according to $K$ value (number of clusters). Each cluster, in turn, will consist of a group of models with similar characteristics in terms of diversity and accuracy. Then we pick up candidates from the resulted clusters based on the following strategies:
\textbf{accuracy first}: we do ascending-rank to the clusters according to their models' accuracy on the pruning set then we choose the top-five candidates from the top-performing cluster, to be deployed on a distributed edge environment.
\textbf{diversity first}: we rank the models in each cluster based on their accuracy on the pruning set, after that we pick up a model with maximum accuracy each cluster then deploy those models to distributed edge environment. The number of candidates will be based on $K$ number of clusters. Results and Analysis section will include accuracy measures for both CIFAR10 and CIFAR100 models following those two strategies.

\subsubsection{deployment to distributed edge environment}
The chosen candidates from the previous step are ready now to be deployed on IIoT devices. We assume that each device holds two or more deep learning models, and applies ensemble learning to combine the hosted models' predictions. Using ensemble learning is essential in our approach as it will enable the devices of providing more accurate decisions, and it will eliminate the potentials of any generalisation issues. However, the process of combining predictions from different models could require more additional computational powers. Thus it's essential to bring more power to the edge devices by attaching hardware accelerators, we use Google Coral that brings machine learning inferences to existing systems.

\subsection{Experimental setup} \label{exp_setup}
\begin{figure}[htbp]
    \centering
    \includegraphics[scale=0.5]{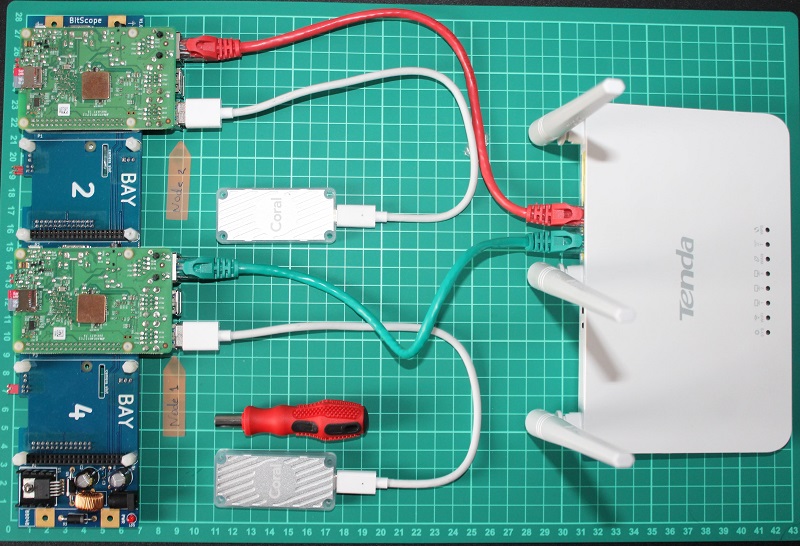}
    \caption{Experimental tools}
    \label{fig:exp_lab}
\end{figure}

the development workstation for training/pruning deep learning models is hosted on Google Cloud Platform with the following specs:8 vCPUs, 30 GB memory. 2 x NVIDIA Tesla K80 using Tensorflow 1.15 dev. For simulating edge computing environment, we use Raspberry PI 3 Model B v1.2 attached to Google Coral (USB accessory that brings machine learning inferencing to existing systems). Google Coral is provided with Google Edge TPU coprocessor, which capable of performing 4 trillion operations (tera-operations) per second. Fig.~\ref{fig:exp_lab} represents the distributed edge environment that we use to evaluate the practicality of our proposed approach. For compatibility reasons with Google Edge TPU device, all models has to be compiled before deployment with a special software (edgetpu\_compiler). Furthermore, to speed up the performance when running multiple models on the edge device, we use use co-compilation. co-compiling allows multiple models to share the Edge TPU RAM to cache their parameter data together, eliminating the need to clear the cache each time you run a different model. Next section will provide the results of co-compilation and normal-compilation.

As shown in this figure, there will be two Raspberry Pi nodes (running "Raspbian GNU/Linux 9 (stretch) and Tensorflow-2.1.0) connected to Tenda F3 300Mbps router. Node1 will host two deep learning models, and Node2 will host three models where all models are generated from CIFAR10 pool.
A master node also is connected to Raspberry Pi nodes to coordinate the predictions process. First It sends to both nodes an integer as the number of required samples that need to be tested on CIFAR10 testing, next the nodes make a prediction on the testing set using ensemble learning and then they send the combined predictions back to the server where the server combines the predictions from both nodes again to produce the final result. while doing the predictions on Node2, we measures the energy consumption (load current,input voltage) using "YOTINO USB Voltage and Current Detector Meter Capacity , Accuracy: ±1\%", and we monitor the temperature of the Node2 using "Etekcity Lasergrip 1080 Non-contact Digital Laser IR Infrared Thermometer, Accuracy: ±2\% or 2°C".

\subsection{Results and discussion}
we present the results after applying \emph{Prune2Edge} on CIFAR10 and CIFAR100 severally , also the results of CIFAR10 includes benchmarking measures for running the generated models on distributed Raspberry Pi devices. Next, we discuss the results and the impact of \emph{Prune2Edge} on edge-ai environment.

\subsubsection{\textbf{results of CIFAR100:}}
\begin{figure}[htbp]
    \centering
    \includegraphics[scale=0.7]{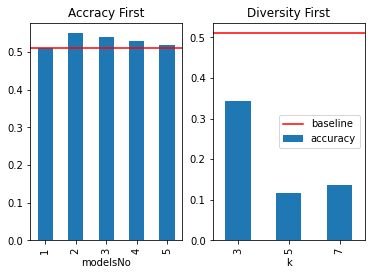}
    \caption{Accuracy of CIFAR100}
    \label{fig:CIFAR100_acc}
\end{figure}

 ResNet model is trained on CIFAR100 data set then pruning is applied to generate a pool of 100 lightweight models. The maximum accuracy in the pruned pool is 66\% while the minimum accuracy is 1\%. Furthermore, the models are compressed, and the size becomes 1.293 KB (almost 63\% smaller than ResNet baseline model). Next, integer-quantisation is applied, so a further reduction in size is gained, and the size becomes 305 KB only, which is a remarkable compression ratio(compression ratio = 92\%). The reduction in the size of the models is a key success factor for our approach toward deploying deep learning models to resource-limited devices. The next step in \emph{Prune2Edge} is to reduce the number of candidates in each pool and pick up the best representatives based on accuracy first and diversity first strategies. Fig.~\ref{fig:CIFAR100_acc} compares the results obtained after adopting the proposed strategies for model selections and apply ensemble learning on the candidates. The accuracy is tested over five hundreds samples on CIFAR100 testing set, and the baseline accuracy on this subset (after quantisation) is 51\%.
 
 From Fig.~\ref{fig:CIFAR100_acc} we can see that the synthesis of the best candidates following "Accuracy First" strategy was able to achieve higher predictability levels than the baseline model. The highest accuracy is achieved when the predictions of the two top models are synthesised using max voting (55\%). On the other hand, the accuracy was remarkably dropped when "Diversity First" is adopted with $k=3,5,7$. When $k=3$, three models (one per cluster) are synthesised with 34\% accuracy only.

\subsubsection{results of CIFAR10:}
\begin{figure}[htbp]
    \centering
    \includegraphics[scale=0.7]{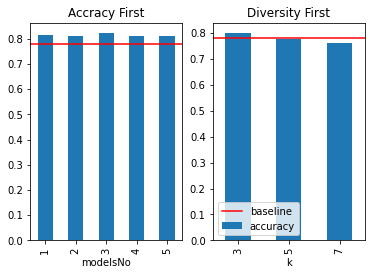}
    \caption{Accuracy of CIFAR10}
    \label{fig:CIFAR10_acc}
\end{figure}

The minimum accuracy in the pruned pool of CIFAR10 is 1\% and the maximum is 80\% (on the training set). The size of the models is 4,926 KB (around 48\% reduction in models size compared to the baseline). After quantization, the models are become even smaller and the size becomes 1,233 KB (around 87\% reduction in size) which easily allow to run those models on resource-limited environment. In similar manner to CIFAR100, Fig.~\ref{fig:CIFAR10_acc} presents the accuracy of \emph{Prune2Edge} on five hundreds image from CIFAR10 testing set, and the accuracy of the baseline model is on this subset reaches 78\%.
What stands out in Fig.~\ref{fig:CIFAR10_acc} is the success of our approach to providing more accurate decisions than the baseline model in both selection strategies. In the case of "Accuracy First", all the ensembles are able to provide outperforming results, and the maximum accuracy is reached when ensemble size is three (82\%). In "Diversity First" approach, when $k=3$ the accuracy of the composed ensemble is 79\%. However, the accuracy was almost the same as the baseline when the ensemble size is five $k=5$ (77\%), and the ensemble size is seven $k=7$ (76\%).
So far, the results indicate that the \emph{Prune2Edge} can provide lightweight and outperforming models in terms of diversity, size and accuracy. Moving now to provide benchmarking results related to heating, inference time and energy consumption on Raspberry Pis.
As explained earlier in section \ref{exp_setup}, we compile the output of \emph{Prune2Edge} using edge\_tpu compiler (co and regular compilation) then we deploy the models to distributed Raspberry Pi devices. Both of Table.~\ref{table:cocompilation} and Table.~\ref{table:nocompilation} show the results of inference time (seconds) on a different number of CIFAR10 testing set, in addition to the external temperature of Node2 (holds three deep learning models) in the rest and the peak (making an inference).

\begin{table}[ht]
\caption{Inference and Temperature Results, Co-Compilation}
\centering
\begin{tabular}{|c|c|c|c|c|c} 
\hline 
Samples & Node1 & Node2 & Temp\_idle& Temp\_Peak \\ 
\hline 
\textbf{100} & 4.242 & 4.562 &  35.9 & 36.2  \\ 
\hline
\textbf{50} & 3.790 & 4.018 & 36 & 36.1\\\hline
25 & 3.638 & 3.818 & 36 & 36.1\\\hline
10 & 3.518 & 3.612 & 35.9 & 36\\\hline
5 & 3.506 & 3.597 & 35.8 & 35.8 \\\hline
4 & 3.494 & 3.563 & 36.1 & 36.1 \\\hline
3 & 3.492 & 3.580 & 35.8 & 35.8 \\\hline
2 & 3.509 & 3.575 & 36.1 & 36.1 \\\hline
1 & 3.503 & 3.546 & 36 & 36 \\ 
\hline 
\end{tabular}
\label{table:cocompilation} 
\end{table}

\begin{figure}[htbp]
    \centering
    \includegraphics[scale=0.6]{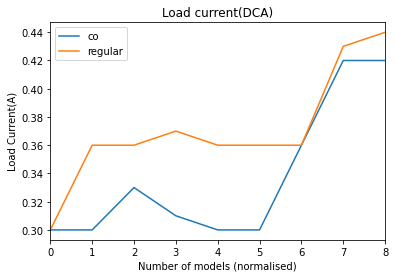}
    \caption{Energy Consumption}
    \label{fig:energy}
\end{figure}

\begin{table}[ht]
\caption{Inference and Temperature Results, Regular-Compilation}
\centering
\begin{tabular}{|c|c|c|c|c|c} 
\hline 
Samples & Node1 & Node2 & Temp\_Idle & Temp\_Peak \\ 
\hline 
\textbf{100} & 11.504 & 15.476 & 36.1& 36.5  \\\hline
\textbf{50} & 7.605 & 9.544 & 36.1 & 36.4\\\hline
25 & 5.473 & 6.421 & 36.1 & 36.2\\\hline
10 & 4.196 & 4.551 & 36.1 & 36.1\\\hline
5 & 3.756 & 3.954 & 36.1 & 36.1 \\\hline
4 & 3.693 & 3.874 & 36 & 36 \\\hline
3 & 3.631 & 3.769 & 35.9 & 35.9 \\\hline
2 & 3.548 & 3.658 & 36 & 36 \\\hline
1 & 3.548 & 3.541 & 36 & 36 \\ 
\hline 
\end{tabular}
\label{table:nocompilation} 
\end{table}
Closer inception for the tables above shows that both compilation methods have almost similar results when the number of samples is between [1-5]. The external temperature of Node2 has not changed (before/after inferencing) and the average time needed to perform an inference is 3.5 seconds. However, when the number of samples is increased, co-compilation shows a significant difference in performance. For instance, when we ask Node2 to make an inference on a hundred images, it takes only 4.56 seconds. On the other hand, doing the same when the models are compiled using regular-compilation it takes almost 15.5 seconds. Additionally, the external temperature of Node2 is slightly increased when it tries to make a prediction on 100 images (0.4 degrees). In short, we could say that, if the devices need to perform inferencing on a large number of samples then the models should be compiled with the co-compilation approach as this benefits from using models parameters cashing and brings essential enhancement for the performance.
With regards to energy consumption, Fig.~\ref{fig:energy} displays the changes of the "load current" while Note2 makes inferencing on different number of samples from CIFAR10 testing set.

From the chart above, it can be seen that Node2 pulls (0.3 A) in rest mode. While doing the inferencing (using co-complied models and the number of samples is less than fifty) the "load current" is slightly changed. The "load current" is increased by almost 0.12, when Node2 tries to make predictions on fifty and a hundred images. When inferencing is done with regular-complied models, the current is slightly increased even when the number of testing samples is less than fifty. It worth to mention that in both compilation methods, the input voltage for Node2 during the inference has not changed (5.04 V).  This indicates that co-compilation has not only impact on the inferencing times, but also helps to reduce the required energy to run ensembles on resource-limited devices.

\subsubsection{discussion}
The results of this study show that "Edge2Prune" is able to provide deep learning models that could work smoothly on resource-limited devices without draining its resources. The proposed approach applies multi-phase pruning pipelines to reach high compression ratios (up to 91\%), and better accuracy levels. In addition to our aim in providing outperforming models, we are particularly interested in real-world applications of "Edge2Prune". For instance, the output of \emph{Prune2Edge} could be deployed on an inspection robot used in industrial sectors to discover defects. The robot could take a picture for the product on a production line, then send the captured photo to "Edge2Prune" ensembles to make accurate predictions about the defects.
The results of Raspberry PIs show the efficiency of this approach in preserving the resources of the device. However, was a Pi device needs in average 3.5 seconds to make an inference, and this could be a critical factor in some real-world applications. Thus, there are very promising potentials for further investigations that include: investigating different combat-baseline architectures that are carefully designed to run on resource-limited environments which could lead to significant improvements on the inference time and even the size of the models.

\section{Conclusion and Future work}
\label{conclusion}
In this paper, we present \emph{Prune2Edge} multi-phase pruning pipelines to deep ensemble learning on resource-constraint devices. It leads to improved classifiers in terms of generalisation, size, accuracy and inferencing on AI edge endpoints that include TPUs or CPUs. First, we generate a diverse-pruned pool of deep learning classifiers, then integer-quantisation is applied to the pool. After that, we shrink the size of the pool through a new clustering-based ensemble pruning technique. Next, the remaining learners after ensemble pruning are optimised then deployed to IoT devices to compose ensembles of deep learning models. 
The experiments results on CIFAR10,CIFAR100 are promising, the compression ratio reaches 91\%, and the composed ensembles are able to beat the predictability levels of a baseline model. Furthermore, the simulation on the distributed Raspberry Pi devices shows the ability of \emph{Prune2Edge} to preserve the resources of the device and runs ensemble learning smoothly. At present we, we count on very popular CNN architectures like ResNet that are not optimised for the resource-limited environment, and we use a simple ensemble learning technique called max voting. In the future, we aim to rely on combat CNNs architectures that are uniquely designed to run on mobiles and IoT devices like MobileNet3, in addition, we could use more advanced ensemble learning techniques on the edge devices like boosting and bagging then we test our approach on more challenging datasets like ImageNet. An exciting application for this future work could be very beneficial in IIoT environments like industrial inception robots.

\section*{Acknowledgement}

The authors would like to thank Google for the generous credit grant that allows using cutting-edge deep learning frameworks on their cloud platform.

\ifCLASSOPTIONcaptionsoff
  \newpage
\fi

\bibliographystyle{IEEEtran}
\bibliography{references}

\end{document}